\documentclass[
]{ceurart}

\newcommand\norm[1]{}
\begin{document}
%%

\copyrightyear{2023}
\copyrightclause{Copyright for this paper by its authors.  Use permitted under Creative Commons License Attribution 4.0 International (CC BY 4.0)}

\conference{NORMalize 2023: The First Workshop on the Normative Design and Evaluation of Recommender Systems, September 19, 2023, co-located with the ACM Conference on Recommender Systems 2023 (RecSys 2023), Singapore}

\title{Improving and Evaluating the Detection of Fragmentation in News Recommendations with the Clustering of News Story Chains}

\author[1]{Alessandra Polimeno}[%
,
email=a.a.polimeno@uu.nl,
url=,
]
\author[2]{Myrthe Reuver}[%
,
email=myrthe.reuver@vu.nl,
orcid=https://orcid.org/0000-0002-5676-0872,
]
\author[3]{Sanne Vrijenhoek}[%
,
email=s.vrijenhoek@uva.nl,
url=,
]
\author[2]{Antske Fokkens}[%
,
email=antske.fokkens@vu.nl,
orcid=https://orcid.org/0000-0002-6628-6916,
]

\address[1]{Utrecht Data School, Utrecht  University, the Netherlands}
\address[2]{Computational Linguistics \& Text Mining Lab, Vrije Universiteit Amsterdam, the Netherlands}
\address[3]{Institute for Information Law, University of Amsterdam, the Netherlands}

\begin{abstract}
News recommender systems play an increasingly influential role in shaping information access within democratic societies. However, tailoring recommendations to users' specific interests can result in the divergence of information streams. Fragmented access to information poses challenges to the integrity of the public sphere, thereby influencing democracy and public discourse. The Fragmentation metric quantifies the degree of fragmentation of information streams in news recommendations. Accurate measurement of this metric requires the application of Natural Language Processing (NLP) to identify distinct news events, stories, or timelines. This paper presents an extensive investigation of various approaches for quantifying Fragmentation in news recommendations. These approaches are evaluated both intrinsically, by measuring performance on news story clustering, and extrinsically, by assessing the Fragmentation scores of different simulated news recommender scenarios. Our findings demonstrate that agglomerative hierarchical clustering coupled with SentenceBERT text representation is substantially better at detecting Fragmentation than earlier implementations. Additionally, the analysis of simulated scenarios yields valuable insights and recommendations for stakeholders concerning the measurement and interpretation of Fragmentation. 
\end{abstract}


\maketitle

\pagenumbering{arabic}
\section{Introduction}
In recent years, the public debate has become increasingly polarized and fragmented~\cite{newman2022reuters,spohr2017fake,hart2020politicization}. News recommender systems may have a role to play in alleviating both these issues by helping users navigate the available content, and assisting them in finding those articles that are interesting or relevant to them. This may increase enjoyment of and engagement with the news, while exposing them to viewpoints or ideas different from their own may increase tolerance and understanding \cite{mutz2002cross, munson2013encouraging, mattis2022nudging}. To facilitate the discussion on the amount of shared experiences between users within a recommender system, \citet{vrijenhoek2021recommenders} proposes the \textit{Fragmentation} metric, which will be explained in more detail in Section~\ref{sec:FragDef}.  

For Fragmentation, merely measuring user engagement with individual articles falls short of capturing the desired granularity. Instead, the metric should focus on aggregating articles into larger units, such as news events or stories. This paper presents an implementation and evaluation of news story detection methods in a recommendation set. We highlight the advantages, limitations, and implications of different approaches for measuring Fragmentation. These approaches range from a basic TF-IDF and cosine similarity-based clustering approach used by \citet{vrijenhoek2021recommenders} to an implementation based on SentenceBERT (SBERT) \cite{reimers-2019-sentence-bert} with agglomerative hierarchical clustering. Our intrinsic evaluation uses a dataset of news articles with gold standard human labeling for news stories \cite{laban2021NewsHG}. Intrinsic evaluation compares clustering approaches to gold standard labels. We also perform extrinsic evaluation: we evaluate our approaches on the capability to detect diverging Fragmentation in (simulated) user scenarios. Our findings show that a SentenceBERT approach with agglomerative hierarchical clustering considerably improves the capability to detect Fragmentation in both intrinsic and extrinsic evaluations. Capturing low Fragmentation remains challenging, which demonstrates the importance of adding extrinsic evaluation - whether in real or simulated recommendation scenarios. We also discuss practical implications for stakeholders and developers aiming to implement Fragmentation metrics. This study represents the first extensive evaluation of Fragmentation quantification that uses intrinsic and extrinsic evaluation, and provides actionable interpretation recommendations.\footnote{An earlier version of this work can be found in \citet{Polimeno}.}

\section{Fragmentation: From Theory to Operationalizaion}

This section introduces and contextualizes the concept of Fragmentation in news recommendation, discusses related work in NLP, and proposes the operationalization approaches implemented in this paper. 

\subsection{Fragmentation in News Recommendation
and Theories of Democracy}\label{sec:FragDef} 

News recommender systems have 
a significant role in shaping the exposure of news articles to the wider public, effectively replacing the traditional gatekeeping function previously entrusted to editorial teams. The editorial teams would ensure that the combination of items featured on a newspaper's front page, or later the landing page of a website or app, would align with the organization's norms and values. Designers of recommendation systems must consider their implicit and explicit design choices, as they can have profound implications for informational awareness and public discourse. Understanding the diverse objectives and editorial standards of news recommender systems is essential, as they vary depending on the desired outcomes. \citet{helberger2019democratic} outlines four distinct societal roles or models of news recommender systems, drawing from the democratic normative framework. \citet{vrijenhoek2021recommenders} proposes five news diversity metrics based on this foundation, which are further refined in \citet{vrijenhoek2022radio}).

The goal of \textit{Fragmentation}, one of the metrics proposed in \citet{vrijenhoek2021recommenders}, is to assess the level of a common public sphere among users within the system: if the Fragmentation score is high, there is little overlap between the recommendations issued to different users. In contrast, users see much of the same content when Fragmentation is low. Fragmentation is mostly relevant to the Liberal and Participatory models of democracy described by \citet{helberger2019democratic}. The Liberal model assumes that ultimately the individual user knows what is best for them and personal preferences are deemed highly important. This model thus allows for a high Fragmentation score. In contrast, the Participatory model aims to give users the information they need in order to play an active role in society. Users should be informed of important events, but in a manner that is suitable for their individual circumstances and capabilities. This touches on an important aspect of Fragmentation: it should not capture whether users read exactly the same articles, but whether they read about the same events or \emph{stories}. The Participatory model of democracy explicitly aims for a lower Fragmentation score, indicating a more homogeneous public sphere.
Fragmentation is defined as a between-user metric, specifically representing differences in recommended \textit{news story chains}, or sets of articles describing the same issue or event from different perspectives. Computationally, Fragmentation is calculated with 1-Rank Biased Overlap as used in \citet{webber2010similarity}, with 1 denoting two completely disjoint recommendation lists and 0 a perfect match. 
The aggregate Fragmentation score of a news recommender is the average Fragmentation score over all pairs in the set of users.

\subsection{Measuring Fragmentation and Diversity with NLP}\label{sec:NLPforFrag}

Fragmentation requires a method to detect stories, timelines, or ideas mentioned in news texts, in order to measure differences between them in recommendation. There are several relevant Natural Language Processing (NLP) tasks that can be used to tackle this. We follow \citet{reuver2021no} and look beyond one NLP task definition, reviewing literature on several related NLP tasks: \textit{news story chain detection}, \textit{news event detection}, \textit{viewpoint detection}, and in general the detection of difference and similarity in news texts. 

\citet{vrijenhoek2021recommenders} mentions \textit{news story chain detection} as relevant for Fragmentation. News story chains are sets of articles that report on the same action, event, or incident \citep{nicholls2019understanding}, such as a presidential speech, a match in a sports game, or a specific protest. \citet{nicholls2019understanding} address this task using a three-day moving window, since most story chains are relevant for three days or less.
News chain detection is challenging, because new topics and events appear continuously, and societal relevance can change on a daily basis. Supervised machine learning methods require labelled data and are not suitable for dealing with a scenario where completely new data appears at a regular basis. Therefore, unsupervised methods are considered more suitable for this task\cite{vrijenhoek2021recommenders, trilling2020between}. Story chains can be detected by calculating pair-wise similarity scores for all articles in a news article corpus, and subsequently generating clusters with a bag of words TF-IDF \cite{sparck1972statistical} text representation to obtain the news story chains \cite{webber2010similarity, boumans2018agency, nicholls2019understanding}. \citet{trilling2020between} use a Word2Vec \cite{mikolov2013efficient} model pre-trained on Dutch news and apply graph-based and hierarchical clustering to identify news story chains. They find no significant difference between the two clustering methods. 

\citet{mulder2021operationalizing} aim to detect content difference in news recommendations, and operationalize this with Entman Frames \cite{entman1993framing} to measure different versus similar framings of the same issues in the news. Their evaluation of the recommendation diversity uses Intra-List Diversity, calculating the average distance between pairs of unique articles. Distance was determined with twenty pre-defined topics (cosine distance) and LDA topic model distributions (Kullblack-Leibler Divergence).  

\citet{hada2023beyond} operationalize Fragmentation as the overlap between social media users' exposure to different viewpoints, specifically on the topic of immigration. A viewpoint is defined as whether it expresses a \textit{diagnostic claim}: the explicit mention of a  problem with a topic. These diagnostic claims are detected in a supervised manner by fine-tuning a BERTweet model. Fragmentation is then measured as overlap exposure to different viewpoints at the pair-wise level, and specifically similarity between user vectors of interactions with viewpoints. \citet{alam2022towards} operationalize ``different viewpoints" in news recommendations with stance and sentiment towards the topic of immigration, and calculate difference over both individual users and the recommender system as a whole. In this manner they determine the diversity (and fragmentation) of the users' reading behaviour.

\subsection{Considerations on Operationalization}\label{sec:Opera}

Operationalization is the process of developing specific methods or approaches to empirically and concretely measure a complex construct \cite{mueller2004}. Since Fragmentation is a complex construct, it becomes especially important to determine when such an operationalization is successful. \citet{jacobs2021measurement} introduce practices on measurement modelling and construct validity from the social sciences to computer science research. They argue computational operationalization of a complex construct should involve writing out assumptions on the measurement, and should not only be evaluated on \textit{predictive validity} (e.g.\ the common computer science approach of how well the approach can predict labels in an unseen test set), but should also involve testing the broader notion of \textit{construct validity}. Predictive validity is only one aspect of construct validity: others are \textit{hypothesis validity} (does the measure support well-known hypotheses on the construct?) and \textit{convergent validity} (do different operationalization reveal the same (expected) measurement results?). Evaluation in such a manner is not a binary yes/no answer to a question, but rather a careful consideration. Assumptions and conditions for successful operationalization in the specific use context need to be explicitly stated. In our experiments, we aim to investigate not only predictive validity, but also  investigate whether we actually measure the construct Fragmentation. We do this not only by evaluating whether our approach can predict labels of unseen test examples (\textit{predictive validity}), but also by testing whether we can measure Fragmentation in generated scenarios. In these scenarios, we know the expected level of Fragmentation our approaches should detect. This evaluation in scenarios constitutes both \textit{hypothesis validity} and \textit{convergent validity}.

\section{Experiments}

We aim to detect news story \textit{timelines} to measure Fragmentation in news recommendation sets. Our experiments focus on text representations and clustering algorithms, which we evaluate intrinsically as well as extrinsically.
The code of our experiments can be found in our repository.\footnote{ \href{https://github.com/myrthereuver/clustering_fragmentation/}{\color{blue}{https://github.com/myrthereuver/clustering\_fragmentation/}}.}

\subsection{Data}\label{sec:Data}
The HeadLine Grouping Dataset (HLGD) \cite{laban2021NewsHG} is an American English dataset of news headlines and groupings of headlines on the same events. This dataset refers to the groupings as \textit{timelines}, and define this as articles ``describing the same event: an action that occurred at a specific time and place". \cite[p. 3187]{laban2021NewsHG}. \citet{laban2021NewsHG}'s three authors as well as eight crowdworker annotators determined in a pair-wise fashion whether two news headlines belong to the same or a different timeline. The average agreement, calculated in Adjusted Mutual Information, between smaller event clusters and a leave-one-out approach is .81. 
In total, this dataset contains 1.679 news articles from 34 different news sources, and the annotation procedure led to 10 timelines on the same large-scale event, ranging in size from 80 to 274 news articles.
These timelines are similar to the news story chains used in \citet{vrijenhoek2022radio} and the HLGD has reliable, high-quality human annotations. These characteristics make the dataset suitable to evaluate our NLP approaches on detecting different news stories. 

We use the ten broader granularity labels of timelines, e.g.\ timelines called ``WikiLeaks Trials" or ``Ivory Coast Army Mutiny" - instead of the more fine-grained timeline labels on sub-events. These high-level event labels clearly relate to Fragmentation: users unaware of one of these events are potentially not inhabiting a shared collective information sphere. Events of lower granularity may show results that are more difficult to interpret as Fragmentation, as these sub-events are still somehow related to one main event. However, the small set of 10 larger scale timelines in this dataset does mean we have a comparably small dataset for our experiments. This may be a caveat for our findings.
\citet{laban2021NewsHG} included the URLs in their dataset, allowing us to extract the original text through web scraping. Articles that were inaccessible due to paywalls, removed since the dataset's publication, or contained unreadable text (e.g.\ HTML-only content) were excluded from our analysis. We eliminated recommendations to other articles found at the end of the text, since this may introduce noise. Our final dataset comprises 1,394 news articles representing 10 distinct news timeline stories, from the topic of Human Cloning to the Brazil Dam Disaster.

As we employed unsupervised approaches, there is no dedicated training set. The development set, used to prevent overfitting on specific stories for development and hyperparameter tuning, consists of 363 articles encompassing timelines 1, 2, and 4. We chose these timelines because they are diverse in terms of content and event cluster size,  see the first column of Table~\ref{tab:error_counts} for group sizes and topic content. The evaluation set contains the 1,031 news articles from the remaining 7 story chains.

\subsection{Representation of News Texts}\label{sec:represent}
Before we can automatically detect timelines or cluster texts, we have to represent the articles in a numerical (e.g.\ machine-readable) manner. The choice of text representation plays a crucial role in the clustering process, as it determines how the similarity between articles is measured. We investigate three distinct text representation strategies, each associated with its own limitations.

\textbf{Bag of Words/TF-IDF} News articles are represented in a one-hot matrix encoding indicating vocabulary words, following \citet{vrijenhoek2021recommenders}. All words are lowercase and punctuation is removed, as well as all stopwords using scikit-learn \cite{pedregosa2011scikit}. \norm{With this sparse matrix, TF-IDF calculations can be used to find words that uniquely contribute to distinguishing different documents.} This representation loses information on word order and also results in a sparse representation. Our TF-IDF baseline approach with graph-based clustering also includes filtering for > .5 cosine similarity between text pairs to denote similarity.

\textbf{GloVe Embeddings} GloVe \cite{pennington2014glove} is an unsupervised method for text representation with the basic underlying assumption that more similar words will have a similar frequency and distribution. We use 300-dimenstion pre-trained GloVe embeddings and represent articles by averaging GloVe embeddings. GloVe embeddings capture semantic similarity between words, but are also static and cannot capture context-dependent differences in meaning. 

\textbf{Sentence Embeddings}
Transformer models, such as BERT \cite{devlin2018bert}, are able to learn context-dependent meanings. \norm{enabling them to capture more nuanced semantic representations of the text. Similar to GloVe-based embeddings, these models are trained on large corpora of general language, employing a self-supervised approach that involves pre-training objectives such as predicting masked words or predicting the subsequent sentence in a word sequence. Through this process, the models refine their self-attention mechanisms across multiple layers, enabling them to effectively represent and classify new texts based on their acquired knowledge of general language patterns.}
We represent each article with pre-trained SentenceBERT (SBERT) embeddings \cite{reimers-2019-sentence-bert}. SBERT is specifically designed to capture semantic similarity between documents, which makes it especially useful for our goal of capturing similarity versus difference in news texts in terms of news timelines. SBERT pre-trains an architecture that consists of two identical Transformer networks, with each network receiving a different text as input. The two networks have tied weights, meaning it is able to capture similarity between the inputs. This strategy is called bi-encoding, and leads to more semantically plausible representations.

\subsection{Clustering Methods}\label{sec:clustering}
\textbf{Graph-based clustering} Graph-based Clustering aims to build clusters by representing the data points as a network with nodes and edges - edges within a sub-graph get assigned high weights (which denotes high similarity) while edges across sub-graphs receive low weights (denoting low similarity) \cite{chen2010graph}. There are several downsides to graph-based clustering: it shows high precision, but tends to be conservative in connecting two articles as related, resulting in a lower recall \cite{nicholls2019understanding, trilling2020between}. To construct the graph and determine the edge weights, we employ the Louvain algorithm \cite{blondel2008fast}, following \citet{vrijenhoek2021recommenders}. Their implementation serves as our baseline for comparison.  
 
\textbf{Density-based clustering} Density-based clustering algorithms, such as DBScan \cite{ester1996density}, identify regions of high data density. DBScan operates by identifying \textit{core points}, which are data points surrounded by a minimum number of neighboring data points. Subsequently, clusters are formed by connecting closely located core points. \textit{Border points}, on the other hand, are identified as points that adjoin the nearest cluster and subsequently merge with it. Noise points, which do not exhibit proximity to any cluster, are not considered as part of the clustering process. DBScan has two hyperparameters: \textit{epsilon} (maximum distance between data points to be seen as close to one another) and \textit{minimum samples} - a number of data points required to be a core point. Our hyperparameter tuning systematically tested 29.403 combinations of each parameter setting (1 to 150, in steps of 1) and text representation (3). The best performance, in terms of V measure, turned out to be  32-66 minimum samples for SBERT, 1-99 minimum samples for GloVe representations, and 2 for Bag of Words. 

\textbf{Agglomerative hierarchical clustering} Agglomerative hierarchical clustering (AHC) is a top-down clustering approach wherein each text initially forms its own cluster. Subsequent merging of these clusters is based on their similarity \cite{murtagh2012algorithms}. Two crucial hyperparameters are be considered for optimizing AHC: the distance threshold (or similarity function) and the linkage criterion (which measures the distance between all points). In our study, we conducted a systematic evaluation of four commonly used linkage criteria: Ward's, average linkage, complete linkage, and single linkage. We explored a total of 600 combinations, varying the distance thresholds from 1 to 150 in increments of 1. Among these combinations, Ward's linkage exhibited the best performance in terms of V-measure on the development set. We determined that the ideal distance settings varied based on the specific text representations used, resulting in a distance of 5-9 for SBERT, 4-6 for GloVe and 124-148 for Bag of Words representations.

\subsection{Evaluation}\label{sec:Evaluation}
This section first provides our intrinsic evaluation focused on accurately detecting news story labels. We then provide an extrinsic evaluation that assesses the effectiveness of these methods within a pipeline targeting the detection of Fragmentation differences in simulated recommendation scenarios. 

\subsubsection{Intrinsic Evaluation: Clustering}
Our intrinsic evaluation requires the scoring of the different approaches on their clustering performance. We use five evaluation metrics. The use of multiple evaluation metrics increases the reliability and stability of our findings, since internal validity measures can be highly variable across runs and datasets \cite{arbelaitz2013extensive}. Three metrics, known as external metrics, rely on labeled data to determine cluster membership. \norm{The metrics we use are \textit{Homogeneity}, \textit{Completeness}, and the \textit{V-measure}~\cite{rosenberg2007v} also introduced the.} \textit{Homogeneity} measures the extent to which items within a cluster share the same label. It provides a score ranging from 0 to 1, where a score of~1 indicates a completely homogeneous cluster in terms of labels, while 0 indicates complete diversity of labels. For \textit{Completeness}, a perfect score of 1 indicates that all members belonging to the same class are assigned to a single cluster. Lastly, \textit{V-measure} is the harmonic mean of Homogeneity and Completeness. 
We also incorporate two internal evaluation metrics, namely the \textit{Silhouette Score} and the \textit{Davies-Bouldin Index }(DBI), which do not rely on any labeled data \cite{rendon2011internal}. \citet{rousseeuw1987silhouettes} measure the validity of clusters by calculating how similar each item in a cluster is to other data points in the same cluster and to adjacent clusters. The \textit{Silhouette Score}, ranging from -1 to 1, represents the average calculation of this for all items. \norm{See Formula (1) below, where a(i) is the mean distance between a point and all other points in the same cluster, and b(i) the mean distance between a point and all points in the nearest cluster. We use Euclidean distance for our evaluation.
\begin{equation}
{s(i) = \frac{b(i)-a(i)}{max(b(i)-a(i)}}
\end{equation}\label{eq_sil}}
Higher scores mean that items in the same cluster are more similar to one-another than to items in adjacent clusters, which would indicate better clustering. \norm{The Silhouette Score is easily interpretable, and its interpretation is clearly related to similarity.}
The \textit{Davies-Bouldin Index}, or DBI \cite{davies1979cluster}, compares clusters' average similarity to the closest neighbouring cluster. Similarity is defined as the ratio of distances within a cluster to distances between a cluster. In other words, clusters that are further apart lead to a better score. This measure has 0 as its lowest score, where a lower score indicates better separated clusters. DBI is implemented with Euclidean distance.

\subsubsection{Extrinsic Evaluation: Generating Recommendation Scenarios}\label{sec:eval_extrinsic}
For our extrinsic evaluation, we generate recommendations that are intentionally designed to exhibit different degrees of fragmentation. We simulate 1,000 users who each receive 7 recommendations, a feasible number for calculation yet large enough to simulate different interaction behaviors with news articles. 

\textbf{Scenario 1: Low Fragmentation}
Users in this scenario have a broad interest in all topics recommended to them: their click behavior shows low Fragmentation in the sense that different users have similar behavior. With the ground truth labels from the News Groupings dataset, the users in this scenario will lead to a Fragmentation score of 0. 

\textbf{Scenario 2: High Fragmentation}
This scenario shows users with very specific and fragmented interests. User click behavior is evenly distributed across different story chains. That is: one group of users only reads from one news grouping, and another group only reads from another group. The resulting Fragmentation score is .85 for the ground truth labels from the News Groupings dataset. 

\textbf{Scenario 3: Balanced Fragmentation}
In reality, the extremes of very low and very high fragmented reading behavior will be rare. This scenario simulates a large portion of readers that read articles from multiple groupings, and a smaller portion of simulated users behaving very selectively. We create three user profiles for this scenario. Profile 1 comprises 70\% of the users, who read at least one article from 5 story chains. The 5 story chains are randomly selected per user. To obtain the total number of 7 recommendations per user, we add 2 random articles from 2 of these chains. Profile 2 consists of 15\% of users: they click on 7 articles from 2 randomly sampled story chains (4 articles from chain 1, 3 from chain 2). Profile 3 has the remaining 15\% of users, and shows the low Fragmentation scenario: all users read from each story chain. 

\section{Results}

This section describes the results of two sets of experiments, both intrinsic (predicting ground truth cluster labels) and extrinsic (measuring Fragmentation in recommendation interaction scenarios). As stated in Section~\ref{sec:Data}, all results are based on the 1,031 news articles from the evaluation set, which contains 7 topics unseen during the development phase.
\norm{Section~\ref{sec:intrinsic_eval} outlines the results of the intrinsic evaluation of clustering as an NLP task, and the results of the clustering evaluation are presented in Table~\ref{tab:evaluation_clustering} with error analysis presented in Section~\ref{sec:error-analysis} . Section~\ref{sec:extrinsic} presents the results of the generated recommendation scenarios and measuring Fragmentation, and the extrinsic evaluation in Table~\ref{tab:frag_all_scens}.}

\subsection{Intrinsic evaluation}\label{sec:intrinsic_eval}

We evaluate the performance of the different clustering methods (Graph-based Clustering, Density-Based Clustering, and Hierarchical Clustering) and text representation approaches (Bag of Words, GloVe embeddings and SBERT embeddings) with five evaluation metrics: Homogeneity, Completeness, V-measure, Silhouette Score and DBI. 
The results of the clustering evaluation are presented in Table~\ref{tab:evaluation_clustering}. Our best-performing approach is SBERT*AHC, with a V-measure of .881. In general, hierarchical clustering performed considerably better than the other clustering methods, and the document representation with SBERT outperformed other text representation methods. Almost all approaches outperform the baseline of BoW*Graph-Based Clustering (V = .161) by a large margin, which is the original method used in \citet{vrijenhoek2021recommenders}. GloVe*DB is the exception (V = .004), and does not outperform this baseline. This is mostly due to a noticeably low Homogeneity score. 

\begin{table*}
    \caption{Results of intrinsic evaluation of different clustering methods. Performance metrics: H (Homogeneity), C (Completeness), V (V-measure), S (Silhouette Score), and DBI (Davies-Bouldin Index). The arrows indicates whether a high (upwards arrow) or low (downwards error) score indicates better performance.}
    \label{tab:evaluation_clustering}
    \centering
    \small
    \begin{tabular}{lccccc}
    \toprule
    \textbf{Representation*Clustering} & \textbf{H $\uparrow$} & \textbf{C $\uparrow$} & \textbf{V $\uparrow$} & \textbf{S $\uparrow$} & \textbf{DBI $\downarrow$}  \\
    \midrule
    BoW*Graph-Based (baseline) & 0.166 & 0.156 & 0.161 & -0.060 & 12.441 \\
    \midrule
    SBERT*AHC &\textbf{ 0.921} & \textbf{0.844} & \textbf{0.881} & 0.290 & 1.933  \\
    GloVe*AHC & 0.762 &	0.708 &	0.734 &	0.183 &	\textbf{1.913} \\
    BoW*AHC & 0.813 & 0.658 & 0.727 & \textbf{0.413} & 1.965	\\
    \midrule
    SBERT*DB & 0.694 & \textbf{0.872} & \textbf{0.773 }& 0.231 &	1.509 \\	
    GloVe*DB & 0.002 & 0.236 & 0.004 & \textbf{0.390} & 0.387 \\
    BoW*DB & \textbf{0.993} & 0.283 & 0.441	& 0.213 & \textbf{0.218} \\

    \bottomrule
    \end{tabular}
\end{table*}

The dramatic underperformance of GloVe*DB warrants further investigation. We find that this particular approach generates only three clusters, where two clusters consist of a single notably short article each, while the remaining cluster encompasses all other articles.\footnote{One article displays HTML code instead of the article text, likely due to a mishap in the preprocessing step.} Despite its poor performance, this approach nevertheless yields a high Silhouette Score, indicating a substantial level of similarity among objects within the predicted clusters, while demonstrating dissimilarity from objects in other clusters. This shows that evaluation metrics based on predicted clusters, rather than gold standard labels, can lead to misleading results in assessing the performance of clustering approaches. Another approach that performs relatively poorly is BoW*DB. It scores well on Homogeneity, but low on Completeness. This approach produces a total of 868 clusters, leading to uniform clusters and thus high Homogeneity. However, these many small clusters means that gold labels are spread across several clusters - which leads to low Completeness. 

\textbf{Error Analysis} For more insight into the behavior of each method, we compare errors made by the three best performing approaches: SBERT*AHC, SBERT*DB, and GloVe*AHC. This analysis allows us to both compare approaches with different text representation techniques while having the same clustering approach, as well as approaches with the same text representation but different clustering approaches. The procedure involved a comparative evaluation of predicted labels assigned to individual gold clusters. For each experimental setup, the majority label predicted by the respective approach was determined, and the instances that deviated from this label were recorded. This provides insights into the dissimilarities between the predicted clusters and the gold standard. Its results show whether approaches exhibit similar errors, and reveals which events are challenging to distinguish.

\begin{table*}
    \caption[Misclassified Articles per Gold Cluster] {Misclassified Articles per Gold Cluster. The total error has the number of overlapping errors as SBERT*AHC between brackets. Evaluation consisted of the following news stories: Ireland Abortion Vote (2); Facebook Privacy Scandal (4); Wikileaks Trials (5); Tunisia Protests (6); Ivory Coast Army Mutiny (7); Equifax Breach (8); and Brazil Dam Disaster (9).}
    \label{tab:error_counts}
    \centering
    \begin{tabular}{cc|ccc}
    \toprule
    \textbf{Gold cluster} & \textbf{Size} &\textbf{SBERT*AHC} & \textbf{SBERT*DB} &\textbf{ GloVe*AHC }\\
    
    % & & \multicolumn{3}{l}{\hspace{0.7cm} N errors \textit{(same as SBERT*AHC)}} \\
    \midrule
    2  & 167 & 9 & 5 & 3 \\
    4  & 163 & 4 & 4 & 38 \\
    5 & 152 & 55 & 36 & 54 \\
    6 & 83 & 2 & 2 & 39 \\
    7 & 99 &  20 & 9 & 39 \\
    8  & 126 & 1 & 1 & 20 \\
    9  & 241 & 9 & 1 & 45 \\ 
    Total & & 100 & 58 \textit{(56)} & 212 \textit{(79)} \\ 
    \bottomrule
    \end{tabular}

\end{table*}

Table \ref{tab:error_counts} presents the error counts associated with each investigated setup, as well as the overlap in errors with the best-performing system AHC*SBERT (indicated between brackets after the total error count).  Of the 58 errors that DB*SBERT makes, 56 were also made by AHC*SBERT. AHC*GloVe incorrectly assigns 79 articles that are also assigned to the same cluster by AHC*SBERT. The large overlap indicates that the setups generally find the same articles difficult to cluster correctly, especially both SBERT systems. For all setups, the chain on the Wikileaks Trial seems the most difficult to identify correctly. Manual inspection of the misclassified articles shows that there is apparent confusion with two other news stories - the Facebook Privacy Scandal and the Equifax Breach. These are stories with similar topical content, as they report on a data leak and/or information breach.
We also found that one SBERT*SHC cluster contains all articles from the same news outlet, which are two topically unrelated news events according to the gold labels: the Ivory Coast Army Mutiny and the WikiLeaks trial. Inspection of this cluster reveals the news articles actually contain content on the Russian invasion of Ukraine, a topic that should not have been included in the dataset. Possibly, the news organizations re-used the URLs used for scraping our dataset. Since this appears to be one topic or event, this result indicates that the SBERT*AHC approach is able to identify topically coherent clusters of news timelines and events. 
Glove*AHC presents considerably less coherent clusters. Only the Brazil Dam Disaster and the Ireland Abortion Vote news timelines are identified in a separate and unique cluster. The Equifax Breach event cluster contains several unrelated technology articles. Other clusters are much less topically coherent when using Glove*AHC.

\subsection{Extrinsic evaluation}\label{sec:extrinsic}

Our extrinsic evaluation aims to identify which clustering method provides the best indication of the actual Fragmentation in three user behavior scenarios. Recall that Scenario~1 involves all users reading the same news timelines, which should result in low Fragmentation. Scenario~2 encompasses users reading highly specific timelines, known to lead to high Fragmentation. Lastly, we consider a mixed scenario (Scenario~3) that represents a more realistic user behavior pattern (see Section~\ref{sec:eval_extrinsic}). 
We first calculate Fragmentation in each scenario with the gold cluster labels. Then, we calculate Fragmentation with cluster labels generated by each approach. We also calculate the difference between Scenario~1 and 2 in Fragmentation score to verify whether the two scenarios can be distinguished. Fragmentation scores in these scenarios should be evaluated on the discriminative ability of the metric rather than the specific scores. If the Fragmentation score has hypothesis validity, we expect to be able to discern a higher score from a lower one. The operationalization would be considerably less useful if there is no meaningful difference of Fragmentation score across scenarios. 
Table~\ref{tab:frag_all_scens} shows the results of these experiments. Our findings demonstrate that SBERT*AHC and SBERT*DB display a considerable difference in Fragmentation score across the three scenarios, and are able to distinguish low Fragmentation (Scenario~1) from high Fragmentation (Scenario~2). These approaches also showed superior performance in the intrinsic evaluation, as discussed in Section~\ref{sec:intrinsic_eval}.
SBERT*DB has a lower difference in Fragmentation between Scenario~1 and 2 than SBERT*AHC. This can be explained by the fact that the DB approach generates five clusters, while SBERT*AHC generates nine. Fewer clusters will lead to a lower Fragmentation score, as it increases the likelihood that news articles come from the same timeline. 

\begin{table*}[]
    \centering
    \small
    \caption{Fragmentation score of the different clustering set-ups in different recommendation user behavior scenarios. The direction of the arrow indicates whether a high (upwards arrow) or low (downwards arrow) score is expected based on the user scenario.}
    \begin{tabular}{lcccc}
    \toprule
    \textbf{Representation*Clustering} & \textbf{Scen. 1 $\downarrow$} & \textbf{Scen. 2  $\uparrow$} & \textbf{Scen. 3} & \textbf{Diff 1-2} \\
    \midrule
    Gold Labels & 0.00 & 0.85 & 0.58 & 0.85\\
  BoW*Graph-Based (baseline) & 0.67 & 0.73 & 0.70 & 0.06\\
    \midrule
  SBERT*AHC & 0.31 & 0.87 & 0.64 & 0.56 \\
  GloVe*AHC & 0.38 & 0.84 & 0.63 & 0.46 \\ 
   BoW*AHC & 0.62 & 0.85 & 0.63 & 0.23 \\
    \midrule
    SBERT*DB & 0.16 & 0.74 & 0.48 & 0.58 \\
    GloVe*DB & 0.01 & 0.01 & 0.00 & 0.01\\
    BoW*DB & 0.99 & 0.99 & 0.99 & 0.00 \\
    \bottomrule
    \end{tabular}
    \label{tab:frag_all_scens}
\end{table*}

\section{Discussion}
 
\textbf{Do Extrinsic and Intrinsic Evaluations Align?} In our study, we undertook a comprehensive assessment of story chain clustering as an NLP task, using internal evaluation metrics and metrics that use ground truth labels. \citet{jacobs2021measurement} have pointed out the importance of evaluating beyond predictive validity, that is: beyond predicting out-of-training labels from a test set. Since Fragmentation is inherently connected to user behavior, evaluating whether the operationalization leads to expected results with user behavior is essential. 
How did our evaluation with generated user behaviour align with our intrinsic results?

In general, our results show that intrinsic performance on clustering (evaluated on gold labels) can be a relatively good indicator for the usability of the approach for detecting high Fragmentation. The approaches that scores well on Homogeneity, Completeness and V-Measure were also best at distinguishing differences in Fragmentation when comparing different user scenarios. At the same time, our extrinsic evaluation also showed that even high-performing clustering approaches struggle with scoring user scenarios with low fragmentation, and consistently detect higher Fragmentation than calculations based on gold standard in these scenarios. Cluster internal metrics, on the other hand, proved to be unreliable indicators of the clusters' quality. Clusters that performed poorly on the gold labels in our intrinsic evaluation and showed poor predictors of Fragmentation in our extrinsic evaluation could still have a decent results according to the Silhouette score and the Davies-Bouldin Index. 

These observations provide two important lessons for automatically determining Fragmentation of news. First, systems need to be evaluated using gold truth labels since internal metrics are not sufficiently reliable. Second, it is crucial to employ extrinsic evaluation. We would not have found out that our best clustering methods do not detect extremely low Fragmentation without evaluating the approaches on generated user scenarios. A benefit of the generated user scenarios is that it allows for maximum control of the experimental situation: we know exactly what the expected Fragmentation score is. Precise hypotheses can be constructed, which denote exactly what the approaches should detect in these scenarios. This makes it relatively easy to detect failing operationalizations, and see which scenarios are easy to detect.

\textbf{Considerations for Developers and Stakeholders}\label{sec:disc_stake}
Our results show that the scores calculated based on clustering output cannot be interpreted literally. When there was no Fragmentation, all approaches that differentiated between different scenarios detected a Fragmentation of at least .16. Our overall best-performing approach, differentiating the best between low and high Fragmentation, even detected a Fragmentation score of .31 in the non-Fragmentation scenario. While our approaches did not succeed in the exact measurement of the Fragmentation score reflected by the gold standard labels in a scenario, they were able to reliably distinguish between a high and a low Fragmentation score. Therefore, researchers employing this operationalization of Fragmentation should primarily focus on contrasting interpretations. For instance, it is appropriate to interpret findings such as ``this set of users exhibits significantly lower Fragmentation compared to another set of users.", but not conclude things such as ``this set of users shows low Fragmentation" based solely on a low score obtained from the Fragmentation metric.

\textbf{Limitations} Our Fragmentation approaches have undergone extensive evaluations, but there are limitations inherent in our study. Firstly, the extrinsic evaluation is conducted solely on simulated scenarios. This offers a solid foundation for testing hypothesis validity, but further investigation is required to measure the performance of these approaches in real-world scenarios. Additionally, future work may explore more advanced simulation scenarios, such as agent-based modelling \cite{saga2013evaluating}. Secondly, our intrinsic evaluation experiments are on a set of ten high-level events, rather than on more granular sub-events. The interpretation of Fragmentation becomes more challenging when considering these fine-grained elements, as they may not necessarily represent a fully fragmented information environment. Future work may look into more fine-grained detection of news stories, and their implication for the corresponding Fragmentation scores. An additional reflection is that the number of events is limited, but somewhat topically diverse in types of news events, spanning a military mutiny, a court case (WikiLeaks), and an election event (the Irish abortion vote). Expanding the scope of the intrinsic evaluation is an avenue for future investigation, by evaluating on a higher number of events, different types of events, languages beyond American English, and different cultural contexts. 
Future work could also look into a similar extensive evaluation approach for detecting different perspectives in news articles for the Representation metric \cite{vrijenhoek2021recommenders}.

\section{Conclusion}

Diverging online information environments can lead to a fragmented public sphere, which has implications for public debate and democracy. Consequently, it is crucial for implementers and stakeholders of news recommender systems to explicitly consider the desired level of Fragmentation in recommendations.
This paper systematically examines various approaches for measuring Fragmentation \cite{vrijenhoek2021recommenders}
while operationalizing it as \textit{news story timeline detection}. The experiments cover different text representations (TF-IDF, SBERT) and clustering algorithms (DB-SCAN, graph-based, and agglomerative hierarchical clustering). Our approaches are rigourously evaluated through intrinsic and extrinsic assessments. These experiments show that agglomerative hierarchical clustering, coupled with SentenceBERT text representation, substantially improves the detection of Fragmentation compared to earlier approaches. Notably, our findings highlight semantic text representations with contextualized sentence embeddings (SentenceBERT) is useful for accurately measuring Fragmentation.
Furthermore, we test whether our intrinsic evaluation results on a dataset of news story chains aligns with measuring Fragmentation in an extrinsic evaluation. Simulated user scenarios in this evaluation indicate low Fragmentation scenarios are more difficult to detect than high Fragmentation ones. Our results show improvement over other approaches, but it is important to acknowledge certain caveats: interpretation of Fragmentation necessitates a thoughtful reflection on the societal and individual goals and values of the news recommender system. There is no single optimal or desired level of Fragmentation; rather, a nuanced understanding is crucial. Our presented approach to accurately measure and evaluate Fragmentation serves as a fundamental stepping stone for further investigation into the societal impact of online news environments.

\begin{acknowledgments}
We would like to thank the anonymous reviewers, whose comments helped improve this work. All remaining errors are our own. This research was partially funded through the \textit{Rethinking News Algorithms} project, Open Competition Digitalization Humanities \& Social Science grant 406.D1.19.073 awarded by the Netherlands Organization of Scientific Research (NWO). 
\end{acknowledgments}

%%
%% Define the bibliography file to be used
\bibliography{bibliography}

\begin{thebibliography}{37}
\expandafter\ifx\csname natexlab\endcsname\relax\def\natexlab#1{#1}\fi
\providecommand{\url}[1]{\texttt{#1}}
\providecommand{\href}[2]{#2}
\providecommand{\path}[1]{#1}
\providecommand{\DOIprefix}{doi:}
\providecommand{\ArXivprefix}{arXiv:}
\providecommand{\URLprefix}{URL: }
\providecommand{\Pubmedprefix}{pmid:}
\providecommand{\doi}[1]{\href{http://dx.doi.org/#1}{\path{#1}}}
\providecommand{\Pubmed}[1]{\href{pmid:#1}{\path{#1}}}
\providecommand{\bibinfo}[2]{#2}
\ifx\xfnm\relax \def\xfnm[#1]{\unskip,\space#1}\fi
%Type = Article
\bibitem[{Newman et~al.(2022)Newman, Fletcher, Robertson, Eddy, and
  Nielsen}]{newman2022reuters}
\bibinfo{author}{N.~Newman}, \bibinfo{author}{R.~Fletcher},
  \bibinfo{author}{C.~T. Robertson}, \bibinfo{author}{K.~Eddy},
  \bibinfo{author}{R.~K. Nielsen},
\newblock \bibinfo{title}{Reuters institute digital news report 2022},
\newblock \bibinfo{journal}{Reuters Institute for the study of Journalism}
  (\bibinfo{year}{2022}).
%Type = Article
\bibitem[{Spohr(2017)}]{spohr2017fake}
\bibinfo{author}{D.~Spohr},
\newblock \bibinfo{title}{Fake news and ideological polarization: Filter
  bubbles and selective exposure on social media},
\newblock \bibinfo{journal}{Business information review} \bibinfo{volume}{34}
  (\bibinfo{year}{2017}) \bibinfo{pages}{150--160}.
%Type = Article
\bibitem[{Hart et~al.(2020)Hart, Chinn, and Soroka}]{hart2020politicization}
\bibinfo{author}{P.~S. Hart}, \bibinfo{author}{S.~Chinn},
  \bibinfo{author}{S.~Soroka},
\newblock \bibinfo{title}{Politicization and polarization in covid-19 news
  coverage},
\newblock \bibinfo{journal}{Science communication} \bibinfo{volume}{42}
  (\bibinfo{year}{2020}) \bibinfo{pages}{679--697}.
%Type = Article
\bibitem[{Mutz(2002)}]{mutz2002cross}
\bibinfo{author}{D.~C. Mutz},
\newblock \bibinfo{title}{Cross-cutting social networks: Testing democratic
  theory in practice},
\newblock \bibinfo{journal}{American Political Science Review}
  \bibinfo{volume}{96} (\bibinfo{year}{2002}) \bibinfo{pages}{111--126}.
%Type = Inproceedings
\bibitem[{Munson et~al.(2013)Munson, Lee, and Resnick}]{munson2013encouraging}
\bibinfo{author}{S.~Munson}, \bibinfo{author}{S.~Lee},
  \bibinfo{author}{P.~Resnick},
\newblock \bibinfo{title}{Encouraging reading of diverse political viewpoints
  with a browser widget},
\newblock in: \bibinfo{booktitle}{Proceedings of the international AAAI
  conference on web and social media}, volume~\bibinfo{volume}{7},
  \bibinfo{publisher}{AAAI Press}, \bibinfo{address}{Cambridge, US},
  \bibinfo{year}{2013}, pp. \bibinfo{pages}{419--428}.
%Type = Article
\bibitem[{Mattis et~al.(2022)Mattis, Masur, M{\"o}ller, and van
  Atteveldt}]{mattis2022nudging}
\bibinfo{author}{N.~Mattis}, \bibinfo{author}{P.~Masur},
  \bibinfo{author}{J.~M{\"o}ller}, \bibinfo{author}{W.~van Atteveldt},
\newblock \bibinfo{title}{Nudging towards news diversity: A theoretical
  framework for facilitating diverse news consumption through recommender
  design},
\newblock \bibinfo{journal}{new media \& society}  (\bibinfo{year}{2022})
  \bibinfo{pages}{14614448221104413}.
%Type = Inproceedings
\bibitem[{Vrijenhoek et~al.(2021)Vrijenhoek, Kaya, Metoui, M{\"o}ller, Odijk,
  and Helberger}]{vrijenhoek2021recommenders}
\bibinfo{author}{S.~Vrijenhoek}, \bibinfo{author}{M.~Kaya},
  \bibinfo{author}{N.~Metoui}, \bibinfo{author}{J.~M{\"o}ller},
  \bibinfo{author}{D.~Odijk}, \bibinfo{author}{N.~Helberger},
\newblock \bibinfo{title}{Recommenders with a mission: assessing diversity in
  news recommendations},
\newblock in: \bibinfo{booktitle}{Proceedings of the 2021 Conference on Human
  Information Interaction and Retrieval (CHIIR 21)}, \bibinfo{publisher}{ACM},
  \bibinfo{address}{Canberra, Australia}, \bibinfo{year}{2021}, pp.
  \bibinfo{pages}{173--183}.
%Type = Inproceedings
\bibitem[{Reimers and Gurevych(2019)}]{reimers-2019-sentence-bert}
\bibinfo{author}{N.~Reimers}, \bibinfo{author}{I.~Gurevych},
\newblock \bibinfo{title}{Sentence-{BERT}: Sentence embeddings using {S}iamese
  {BERT}-networks},
\newblock in: \bibinfo{booktitle}{Proceedings of the 2019 Conference on
  Empirical Methods in Natural Language Processing and the 9th International
  Joint Conference on Natural Language Processing (EMNLP-IJCNLP)},
  \bibinfo{publisher}{Association for Computational Linguistics},
  \bibinfo{address}{Hong Kong, China}, \bibinfo{year}{2019}, pp.
  \bibinfo{pages}{3982--3992}. \URLprefix
  \url{https://aclanthology.org/D19-1410}.
%Type = Inproceedings
\bibitem[{Laban et~al.(2021)Laban, Bandarkar, and Hearst}]{laban2021NewsHG}
\bibinfo{author}{P.~Laban}, \bibinfo{author}{L.~Bandarkar},
  \bibinfo{author}{M.~A. Hearst},
\newblock \bibinfo{title}{News headline grouping as a challenging nlu task},
\newblock in: \bibinfo{booktitle}{Proceedings of the 2021 Conference of the
  North American Chapter of the Association for Computational Linguistics:
  Human Language Technologies}, \bibinfo{publisher}{ACL},
  \bibinfo{address}{Online}, \bibinfo{year}{2021}, pp.
  \bibinfo{pages}{3186--3198}.
%Type = Masterthesis
\bibitem[{Polimeno(2022)}]{Polimeno}
\bibinfo{author}{A.~Polimeno}, \bibinfo{title}{Diversifying News Recommendation
  Systems by Detecting Fragmentation in News Story Chains}, Master's thesis,
  Vrije Universiteit Amsterdam, \bibinfo{year}{2022}.
%Type = Article
\bibitem[{Helberger(2019)}]{helberger2019democratic}
\bibinfo{author}{N.~Helberger},
\newblock \bibinfo{title}{On the democratic role of news recommenders},
\newblock \bibinfo{journal}{Digital Journalism} \bibinfo{volume}{7}
  (\bibinfo{year}{2019}) \bibinfo{pages}{993--1012}.
%Type = Inproceedings
\bibitem[{Vrijenhoek et~al.(2022)Vrijenhoek, B{\'e}n{\'e}dict,
  Gutierrez~Granada, Odijk, and De~Rijke}]{vrijenhoek2022radio}
\bibinfo{author}{S.~Vrijenhoek}, \bibinfo{author}{G.~B{\'e}n{\'e}dict},
  \bibinfo{author}{M.~Gutierrez~Granada}, \bibinfo{author}{D.~Odijk},
  \bibinfo{author}{M.~De~Rijke},
\newblock \bibinfo{title}{Radio--rank-aware divergence metrics to measure
  normative diversity in news recommendations},
\newblock in: \bibinfo{booktitle}{Proceedings of the 16th ACM Conference on
  Recommender Systems (RecSys 22)}, \bibinfo{publisher}{ACM},
  \bibinfo{address}{Seattle, USA}, \bibinfo{year}{2022}, pp.
  \bibinfo{pages}{208--219}.
%Type = Article
\bibitem[{Webber et~al.(2010)Webber, Moffat, and Zobel}]{webber2010similarity}
\bibinfo{author}{W.~Webber}, \bibinfo{author}{A.~Moffat},
  \bibinfo{author}{J.~Zobel},
\newblock \bibinfo{title}{A similarity measure for indefinite rankings},
\newblock \bibinfo{journal}{ACM Transactions on Information Systems (TOIS)}
  \bibinfo{volume}{28} (\bibinfo{year}{2010}) \bibinfo{pages}{1--38}.
%Type = Inproceedings
\bibitem[{Reuver et~al.(2021)Reuver, Fokkens, and Verberne}]{reuver2021no}
\bibinfo{author}{M.~Reuver}, \bibinfo{author}{A.~Fokkens},
  \bibinfo{author}{S.~Verberne},
\newblock \bibinfo{title}{No {NLP} task should be an island:
  Multi-disciplinarity for diversity in news recommender systems},
\newblock in: \bibinfo{booktitle}{Proceedings of the EACL Hackashop on News
  Media Content Analysis and Automated Report Generation},
  \bibinfo{publisher}{Association for Computational Linguistics},
  \bibinfo{address}{Online}, \bibinfo{year}{2021}, pp. \bibinfo{pages}{45--55}.
  \URLprefix \url{https://aclanthology.org/2021.hackashop-1.7}.
%Type = Article
\bibitem[{Nicholls and Bright(2019)}]{nicholls2019understanding}
\bibinfo{author}{T.~Nicholls}, \bibinfo{author}{J.~Bright},
\newblock \bibinfo{title}{Understanding news story chains using information
  retrieval and network clustering techniques},
\newblock \bibinfo{journal}{Communication methods and measures}
  \bibinfo{volume}{13} (\bibinfo{year}{2019}) \bibinfo{pages}{43--59}.
%Type = Article
\bibitem[{Trilling and van Hoof(2020)}]{trilling2020between}
\bibinfo{author}{D.~Trilling}, \bibinfo{author}{M.~van Hoof},
\newblock \bibinfo{title}{Between article and topic: News events as level of
  analysis and their computational identification},
\newblock \bibinfo{journal}{Digital Journalism} \bibinfo{volume}{8}
  (\bibinfo{year}{2020}) \bibinfo{pages}{1317--1337}.
%Type = Article
\bibitem[{Sparck~Jones(1972)}]{sparck1972statistical}
\bibinfo{author}{K.~Sparck~Jones},
\newblock \bibinfo{title}{A statistical interpretation of term specificity and
  its application in retrieval},
\newblock \bibinfo{journal}{Journal of documentation} \bibinfo{volume}{28}
  (\bibinfo{year}{1972}) \bibinfo{pages}{11--21}.
%Type = Article
\bibitem[{Boumans et~al.(2018)Boumans, Trilling, Vliegenthart, and
  Boomgaarden}]{boumans2018agency}
\bibinfo{author}{J.~Boumans}, \bibinfo{author}{D.~Trilling},
  \bibinfo{author}{R.~Vliegenthart}, \bibinfo{author}{H.~Boomgaarden},
\newblock \bibinfo{title}{The agency makes the (online) news world go round:
  The impact of news agency content on print and online news},
\newblock \bibinfo{journal}{International Journal of Communication}
  \bibinfo{volume}{12} (\bibinfo{year}{2018}) \bibinfo{pages}{22}.
%Type = Article
\bibitem[{Mikolov et~al.(2013)Mikolov, Chen, Corrado, and
  Dean}]{mikolov2013efficient}
\bibinfo{author}{T.~Mikolov}, \bibinfo{author}{K.~Chen},
  \bibinfo{author}{G.~Corrado}, \bibinfo{author}{J.~Dean},
\newblock \bibinfo{title}{Efficient estimation of word representations in
  vector space},
\newblock \bibinfo{journal}{International Conference on Learning
  Representations}  (\bibinfo{year}{2013}).
%Type = Inproceedings
\bibitem[{Mulder et~al.(2021)Mulder, Inel, Oosterman, and
  Tintarev}]{mulder2021operationalizing}
\bibinfo{author}{M.~Mulder}, \bibinfo{author}{O.~Inel},
  \bibinfo{author}{J.~Oosterman}, \bibinfo{author}{N.~Tintarev},
\newblock \bibinfo{title}{Operationalizing framing to support multiperspective
  recommendations of opinion pieces},
\newblock in: \bibinfo{booktitle}{Proceedings of the 2021 ACM conference on
  fairness, accountability, and transparency}, \bibinfo{publisher}{ACM},
  \bibinfo{address}{Online}, \bibinfo{year}{2021}, pp.
  \bibinfo{pages}{478--488}.
%Type = Article
\bibitem[{Entman(1993)}]{entman1993framing}
\bibinfo{author}{R.~M. Entman},
\newblock \bibinfo{title}{Framing: Toward clarification of a fractured
  paradigm},
\newblock \bibinfo{journal}{Journal of communication} \bibinfo{volume}{43}
  (\bibinfo{year}{1993}) \bibinfo{pages}{51--58}.
%Type = Inproceedings
\bibitem[{Hada et~al.(2023)Hada, Ebrahimi~Fard, Shugars, Bianchi, Rossini,
  Hovy, Tromble, and Tintarev}]{hada2023beyond}
\bibinfo{author}{R.~Hada}, \bibinfo{author}{A.~Ebrahimi~Fard},
  \bibinfo{author}{S.~Shugars}, \bibinfo{author}{F.~Bianchi},
  \bibinfo{author}{P.~Rossini}, \bibinfo{author}{D.~Hovy},
  \bibinfo{author}{R.~Tromble}, \bibinfo{author}{N.~Tintarev},
\newblock \bibinfo{title}{Beyond digital" echo chambers": The role of viewpoint
  diversity in political discussion},
\newblock in: \bibinfo{booktitle}{Proceedings of the Sixteenth ACM
  International Conference on Web Search and Data Mining},
  \bibinfo{publisher}{ACM}, \bibinfo{address}{Singapore}, \bibinfo{year}{2023},
  pp. \bibinfo{pages}{33--41}.
%Type = Inproceedings
\bibitem[{Alam et~al.(2022)Alam, Iana, Grote, Ludwig, M{\"u}ller, and
  Paulheim}]{alam2022towards}
\bibinfo{author}{M.~Alam}, \bibinfo{author}{A.~Iana},
  \bibinfo{author}{A.~Grote}, \bibinfo{author}{K.~Ludwig},
  \bibinfo{author}{P.~M{\"u}ller}, \bibinfo{author}{H.~Paulheim},
\newblock \bibinfo{title}{Towards analyzing the bias of news recommender
  systems using sentiment and stance detection},
\newblock in: \bibinfo{booktitle}{Companion Proceedings of the Web Conference
  2022}, \bibinfo{publisher}{ACM}, \bibinfo{address}{Online},
  \bibinfo{year}{2022}, pp. \bibinfo{pages}{448--457}.
%Type = Incollection
\bibitem[{CW(2004)}]{mueller2004}
\bibinfo{author}{M.~CW},
\newblock \bibinfo{title}{Conceptualization, operationalization, and
  measurement},
\newblock in: \bibinfo{editor}{L.~T. Lewis-Beck~MS, Bryman~A} (Ed.),
  \bibinfo{booktitle}{The SAGE encyclopedia of social science research
  methods}, \bibinfo{publisher}{Sag}, \bibinfo{address}{Beverly Hills},
  \bibinfo{year}{2004}.
%Type = Inproceedings
\bibitem[{Jacobs and Wallach(2021)}]{jacobs2021measurement}
\bibinfo{author}{A.~Z. Jacobs}, \bibinfo{author}{H.~Wallach},
\newblock \bibinfo{title}{Measurement and fairness},
\newblock in: \bibinfo{booktitle}{Proceedings of the 2021 ACM conference on
  fairness, accountability, and transparency (FaccT '21)},
  \bibinfo{publisher}{ACM}, \bibinfo{address}{Online}, \bibinfo{year}{2021},
  pp. \bibinfo{pages}{375--385}.
%Type = Article
\bibitem[{Pedregosa et~al.(2011)Pedregosa, Varoquaux, Gramfort, Michel,
  Thirion, Grisel, Blondel, Prettenhofer, Weiss, Dubourg
  et~al.}]{pedregosa2011scikit}
\bibinfo{author}{F.~Pedregosa}, \bibinfo{author}{G.~Varoquaux},
  \bibinfo{author}{A.~Gramfort}, \bibinfo{author}{V.~Michel},
  \bibinfo{author}{B.~Thirion}, \bibinfo{author}{O.~Grisel},
  \bibinfo{author}{M.~Blondel}, \bibinfo{author}{P.~Prettenhofer},
  \bibinfo{author}{R.~Weiss}, \bibinfo{author}{V.~Dubourg}, et~al.,
\newblock \bibinfo{title}{Scikit-learn: Machine learning in python},
\newblock \bibinfo{journal}{the Journal of machine Learning research}
  \bibinfo{volume}{12} (\bibinfo{year}{2011}) \bibinfo{pages}{2825--2830}.
%Type = Inproceedings
\bibitem[{Pennington et~al.(2014)Pennington, Socher, and
  Manning}]{pennington2014glove}
\bibinfo{author}{J.~Pennington}, \bibinfo{author}{R.~Socher},
  \bibinfo{author}{C.~D. Manning},
\newblock \bibinfo{title}{Glove: Global vectors for word representation},
\newblock in: \bibinfo{booktitle}{Proceedings of the 2014 conference on
  empirical methods in natural language processing (EMNLP)},
  \bibinfo{publisher}{ACL}, \bibinfo{address}{Doha, Qatar},
  \bibinfo{year}{2014}, pp. \bibinfo{pages}{1532--1543}.
%Type = Inproceedings
\bibitem[{Devlin et~al.(2019)Devlin, Chang, Lee, and
  Toutanova}]{devlin2018bert}
\bibinfo{author}{J.~Devlin}, \bibinfo{author}{M.-W. Chang},
  \bibinfo{author}{K.~Lee}, \bibinfo{author}{K.~Toutanova},
\newblock \bibinfo{title}{{BERT}: Pre-training of deep bidirectional
  transformers for language understanding},
\newblock in: \bibinfo{booktitle}{Proceedings of the 2019 Conference of the
  North {A}merican Chapter of the Association for Computational Linguistics:
  Human Language Technologies, Volume 1 (Long and Short Papers)},
  \bibinfo{publisher}{Association for Computational Linguistics},
  \bibinfo{address}{Minneapolis, Minnesota}, \bibinfo{year}{2019}, pp.
  \bibinfo{pages}{4171--4186}. \URLprefix
  \url{https://aclanthology.org/N19-1423}.
  \DOIprefix\doi{10.18653/v1/N19-1423}.
%Type = Inproceedings
\bibitem[{Chen and Ji(2010)}]{chen2010graph}
\bibinfo{author}{Z.~Chen}, \bibinfo{author}{H.~Ji},
\newblock \bibinfo{title}{Graph-based clustering for computational linguistics:
  A survey},
\newblock in: \bibinfo{booktitle}{Proceedings of TextGraphs-5-2010 Workshop on
  Graph-based Methods for Natural Language Processing},
  \bibinfo{publisher}{ACL}, \bibinfo{address}{Upsala,Sweden},
  \bibinfo{year}{2010}, pp. \bibinfo{pages}{1--9}.
%Type = Article
\bibitem[{Blondel et~al.(2008)Blondel, Guillaume, Lambiotte, and
  Lefebvre}]{blondel2008fast}
\bibinfo{author}{V.~D. Blondel}, \bibinfo{author}{J.-L. Guillaume},
  \bibinfo{author}{R.~Lambiotte}, \bibinfo{author}{E.~Lefebvre},
\newblock \bibinfo{title}{Fast unfolding of communities in large networks},
\newblock \bibinfo{journal}{Journal of statistical mechanics: theory and
  experiment}  (\bibinfo{year}{2008}) \bibinfo{pages}{P10008}.
%Type = Article
\bibitem[{Ester et~al.(1996)Ester, Kriegel, Sander, Xu
  et~al.}]{ester1996density}
\bibinfo{author}{M.~Ester}, \bibinfo{author}{H.-P. Kriegel},
  \bibinfo{author}{J.~Sander}, \bibinfo{author}{X.~Xu}, et~al.,
\newblock \bibinfo{title}{A density-based algorithm for discovering clusters in
  large spatial databases with noise},
\newblock \bibinfo{journal}{KDD} \bibinfo{volume}{96} (\bibinfo{year}{1996})
  \bibinfo{pages}{226--231}.
%Type = Article
\bibitem[{Murtagh and Contreras(2012)}]{murtagh2012algorithms}
\bibinfo{author}{F.~Murtagh}, \bibinfo{author}{P.~Contreras},
\newblock \bibinfo{title}{Algorithms for hierarchical clustering: an overview},
\newblock \bibinfo{journal}{Wiley Interdisciplinary Reviews: Data Mining and
  Knowledge Discovery} \bibinfo{volume}{2} (\bibinfo{year}{2012})
  \bibinfo{pages}{86--97}.
%Type = Article
\bibitem[{Arbelaitz et~al.(2013)Arbelaitz, Gurrutxaga, Muguerza, P{\'e}rez, and
  Perona}]{arbelaitz2013extensive}
\bibinfo{author}{O.~Arbelaitz}, \bibinfo{author}{I.~Gurrutxaga},
  \bibinfo{author}{J.~Muguerza}, \bibinfo{author}{J.~M. P{\'e}rez},
  \bibinfo{author}{I.~Perona},
\newblock \bibinfo{title}{An extensive comparative study of cluster validity
  indices},
\newblock \bibinfo{journal}{Pattern recognition} \bibinfo{volume}{46}
  (\bibinfo{year}{2013}) \bibinfo{pages}{243--256}.
%Type = Article
\bibitem[{Rend{\'o}n et~al.(2011)Rend{\'o}n, Abundez, Arizmendi, and
  Quiroz}]{rendon2011internal}
\bibinfo{author}{E.~Rend{\'o}n}, \bibinfo{author}{I.~Abundez},
  \bibinfo{author}{A.~Arizmendi}, \bibinfo{author}{E.~M. Quiroz},
\newblock \bibinfo{title}{Internal versus external cluster validation indexes},
\newblock \bibinfo{journal}{International Journal of computers and
  communications} \bibinfo{volume}{5} (\bibinfo{year}{2011})
  \bibinfo{pages}{27--34}.
%Type = Article
\bibitem[{Rousseeuw(1987)}]{rousseeuw1987silhouettes}
\bibinfo{author}{P.~J. Rousseeuw},
\newblock \bibinfo{title}{Silhouettes: a graphical aid to the interpretation
  and validation of cluster analysis},
\newblock \bibinfo{journal}{Journal of computational and applied mathematics}
  \bibinfo{volume}{20} (\bibinfo{year}{1987}) \bibinfo{pages}{53--65}.
%Type = Article
\bibitem[{Davies and Bouldin(1979)}]{davies1979cluster}
\bibinfo{author}{D.~L. Davies}, \bibinfo{author}{D.~W. Bouldin},
\newblock \bibinfo{title}{A cluster separation measure},
\newblock \bibinfo{journal}{IEEE transactions on pattern analysis and machine
  intelligence} \bibinfo{volume}{PAMI-1} (\bibinfo{year}{1979})
  \bibinfo{pages}{224--227}.
%Type = Inproceedings
\bibitem[{Saga et~al.(2013)Saga, Okamoto, Tsuji, and
  Matsumoto}]{saga2013evaluating}
\bibinfo{author}{R.~Saga}, \bibinfo{author}{K.~Okamoto},
  \bibinfo{author}{H.~Tsuji}, \bibinfo{author}{K.~Matsumoto},
\newblock \bibinfo{title}{Evaluating recommender system using multiagent-based
  simulator: Case study of collaborative filtering simulation},
\newblock in: \bibinfo{booktitle}{Recent Progress in Data Engineering and
  Internet Technology: Volume 1}, \bibinfo{organization}{Springer},
  \bibinfo{year}{2013}, pp. \bibinfo{pages}{155--162}.

\end{thebibliography}

\end{document}